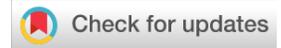

RESEARCH ARTICLE

# AI-generated stories favour stability over change: homogeneity and cultural stereotyping in narratives generated by gpt-4o-mini

[version 1; peer review: awaiting peer review]

Jill Walker Rettberg , Hermann Wigers

Center for Digital Narrative, Department of Linguistic, Literary and Aesthetic Studies, Universitetet i Bergen, Bergen, 5020, Norway

**Open Peer Review**

**Approval Status** AWAITING PEER REVIEW

Any reports and responses or comments on the article can be found at the end of the article.

**Abstract**
Can a language model trained largely on Anglo-American texts generate stories that are culturally relevant to other nationalities? To find out, we generated 11,800 stories - 50 for each of 236 countries – by sending the prompt "Write a 1500 word potential {demonym} story" to OpenAI's model gpt-4o-mini. Although the stories do include surface-level national symbols and themes, they overwhelmingly conform to a single narrative plot structure across countries: a protagonist lives in or returns home to a small town and resolves a minor conflict by reconnecting with tradition and organising community events. Real-world conflicts are sanitised, romance is almost absent, and narrative tension is downplayed in favour of nostalgia and reconciliation. The result is a narrative homogenisation: an AI-generated synthetic imaginary that prioritises stability above change and tradition above growth. We argue that the structural homogeneity of AI-generated narratives constitutes a distinct form of AI bias, a narrative standardisation that should be acknowledged alongside the more familiar representational bias. These findings are relevant to literary studies, narratology, critical AI studies, NLP research, and efforts to improve the cultural alignment of generative AI.

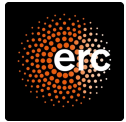 This article is included in the European Research Council (ERC) gateway.

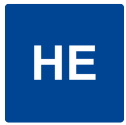 This article is included in the Horizon Europe gateway.

**Corresponding author:** Jill Walker Rettberg (jill.walker.rettberg@uib.no)

**Author roles: Rettberg JW**: Conceptualization, Data Curation, Formal Analysis, Funding Acquisition, Investigation, Methodology, Project Administration, Software, Supervision, Visualization, Writing – Original Draft Preparation, Writing – Review & Editing; **Wigers H**: Data Curation, Formal Analysis, Investigation, Methodology, Software, Writing – Original Draft Preparation, Writing – Review & Editing

**Competing interests:** No competing interests were disclosed.

**Grant information:** This project has received funding from the European Union's Horizon 2020 research and innovation programme under grant agreement number 101142306. The project is also supported by the Center for Digital Narrative, which is funded by the Research Council of Norway through its Centres of Excellence scheme, project number 332643.
*The funders had no role in study design, data collection and analysis, decision to publish, or preparation of the manuscript.*

**Introduction**
What characterises stories generated by large language models like ChatGPT? Large language models, often called generative AI, are being used across the world in industry, the public sector and schools. Versions of ChatGPT, Gemini, and Copilot are becoming integrated into our phones, word processors, text editors, and daily lives, but we still do not know a lot about what characterises the stories generated by large language models (LLMs). The dominant commercial LLMs are American and are largely trained on US data, which has led to concerns about their use in other parts of the world. However, research on cultural alignment and AI bias has largely considered what is known as representational bias at the level of words and images. A well-known example is the depiction of a "terrorist" as Muslim, or a "president" as a white man.

This study explores the hypothesis that a deeper, more formal bias also exists in LLMs, a narrative standardisation at the level of plot structures that might apply narrative structures based on a US context to any story, disregarding local specificity. To imagine how this might impact global cultural diversity think of the differences between a Hollywood movie and a Bollywood movie, or a Latin-American telenovela and a Nordic noir crime drama. If generative AI standardises narratives, and people across the world increasingly use generative AI to help them write speeches, reports, emails and more, we risk losing the cultural diversity that is important to local, regional and national identities.

To test this hypothesis, we generated 11,800 stories using OpenAI's gpt-4o-mini model, which is one of the models accessed by ChatGPT. We used the prompt "Write a 1500 word potential {demonym} story", where {demonym} was a variable, so the model received prompts like "Write a 1500 word potential American story", "Write a 1500 word potential Norwegian story", "Write a 1500 word potential Angolan story" and so on, using 236 countries on Statistics Norway's list of countries. We also generated 50 stories without a demonym to see what the "default" story looked like.

This paper is the first of several analysing the GPT-stories dataset, which is available for download from Dataverse (Rettberg & Wigers, 2025). The paper combined quantitative analyses of stories with qualitative readings of stories from selected countries.

**The standard story: small towns and stability**
Here is an American story, according to OpenAI's generative AI model gpt-4o-mini: A woman returns to her small hometown after spending several stressful years working in the big city. She meets a childhood friend or elderly neighbour who tells her about a problem: the locals are no longer connected to their history, people are leaving, there is a drought, or a developer wants to demolish the old buildings. The protagonist organises a community event and this revitalises the community, allowing them to withstand the challenge. The story concludes with the woman's decision to stay in her hometown and work as an artist/writer/community organiser instead of returning to her busy life in the big city.

This is the plot of most of the 50 stories we generated by sending the prompt "Write a 1500 word potential American story" to OpenAI's large language model gpt-4o-mini. There are a few exceptions. Sometimes the protagonist is a child or a man or a grandmother. Sometimes the protagonist collects people's stories or organises community gardening or re-opens the abandonned train station and makes it a community hub instead of organising a festival. But the stories always involve a protagonist revitalising small town America by engaging the community to be proud of their heritage.

European stories follow the same basic plot structure, but often with supernatural elements. In Norwegian stories, for instance, local girls in small villages who have often recently returned home from the big city wander into forests and meet spirits who tell them that the village/nature/their community is threatened by an imbalance between people and nature, or by developers, or by approaching extreme weather. In most of the stories, the protagonist resolves this by organising the community as they did in the American stories.

In a story generated for Ghana, the elders in the village have grown weary and the young people have forgotten the traditions because they are "absorbed in smartphones and the lure of the city" (this story has the ID GH_1 in our dataset), so the protagonist organises a festival and teaches everyone traditional dances, thus revitalising the village.

You may begin to see a pattern here. A protagonist in a small town or village organises a festival that revitalises a community by rekindling their heritage. There are some exceptions. In some of the Indian stories, for example, the protagonist's goal isn't to heal their village but to leave it to get an education (IN_1) – yet in many of these the protagonist returns (IN_2), bringing us back to the default storyline.

All the stories include symbols and clichés that are commonly associated with the countries they are generated for: mountains and fjords for Norway, olive trees for Israel and Palestine. This sort of association is what LLMs excel at: they model the training data by identifying words and concepts that tend to co-occur, like *fjord* and *Norway*. These co-occurrences or associations are modelled as vectors (imagine an arrow pointing from *fjord* to *Norway*) in what is called the model's semantic space, latent space or vector space. The model has billions of vectors across billions of dimensions, so *fjord* points not only to *Norway* but also to *water* and *deep*. If most of the training data is American, the model will probably also associate *fjord* with *Elsa* and *Anna* from the movie *Frozen* and with *cruise*, due to all the websites marketing fjord cruises to US tourists, whereas a model trained on the local newspaper in a town by the fjord would be unlikely to mention the Disney movie, but might associate *fjord* with *transport, bridge, ferry, tourists, factory* and *fishing*.

These obvious stereotypes, which we will discuss in more detail below, confirm our assumption that OpenAI's model's associations with most countries see the countries *from outside*: they associate fjords with *Frozen*, not with bridges or salmon farming[1]. This relates to research on AI bias and cultural alignment of AI models and shows how a model being sold to countries all over the world is not currently truly international (Adilazuarda *et al.*, 2024; Agarwal *et al.*, 2025; Srdarov & Leaver, 2024; Tao *et al.*, 2024).

What interests us most in this paper, however, is a deeper level of AI bias: a bias at the level of plot and narrative structure. Our dataset shows that generative AI defaults to a very specific plot structure where the resolution is achieved by the protagonist restoring lost traditions and community in small towns and villages. Human-authored stories are far more diverse, although, as we will return to, corporate storytelling genres like Hallmark movies have for many years generated very homogenous stories even before AI.

Language models are *normalising*: they generate content based on statistical probabilities, which means that outliers are excluded and common patterns are amplified (JW Rettberg, 2023, 118–23). A recent study found that generative AI tools in our writing environments homogenised human writing, and that Indian participants' writing became closer to Western norms when they used AI-assisted writing tools than when they wrote without them (Agarwal *et al.*, 2025). What does this mean for how people around the world can express and develop their cultural identities in a digitised world? If AI-generated stories normalise and homogenise our ability to express ourselves, we risk losing the diversity of our cultural heritage.

This paper is the first from a research project funded by the European Research Council that seeks to understand how narrative archetypes in the training data structure the stories LLMs are able to tell[2]. We use the word *archetype* broadly here to include all aspects of narratives that are not at the surface level of word choice and style. We are interested in features ranging from the basic requirement that something must happen in a narrative, to recurring characters like the villain, the love interest and the trickster, to genre-specific archetypes such as being given three wishes, the villain dying at the end of the story, or happy endings with a wedding and happily ever after. In this study, we found that gpt-4o-mini flattens out differences and generates stories with one basic plot structure. There is surface-level stylistic variation from country to country, but using symbols and clichés that are not truly local to the countries they represent.

In the rest of this paper, we first outline some of the theoretical background for our study, looking briefly at the structuralist narratology of the 20th century, contemporary theories of narrative as a rhetorical interaction between the storyteller and the audience, computational narrative generation and emerging theories of narrative and AI, including Bajohr's term *surface narration*. Then we describe what we know about the training data used for OpenAI's GPT models and briefly discuss AI bias and the notion of a *synthetic imaginary* followed by a section on culturally specific narratives.

Next, we describe our methodology for generating the dataset of AI-generated stories. The dataset itself is published on Dataverse (Rettberg & Wigers, 2025), and the scripts used to generate the dataset and scripts used for analysing it are available on GitHub and archived on Zenodo (Wigers & Rettberg, 2025).

After explaining the methodology we analyse the stories, first using word frequency analysis to obtain an overview, then using word tree analysis and close reading to compare the use of olive trees and words like *stand* and *fight* in Palestinian and Israeli stories. We move on to a closer reading of the striking use of trains as a symbol in the American stories and finally provide a qualitative analysis of the Norwegian stories, drawing out the plot structure. Our conclusion explains why this topic matters and describes further research that is needed.

**AI bias and narrative structure**
Stories are powerful because they express a specific person's experience in a specific situation. They express what Donna Haraway calls *situated knowledge* (1988), a way of understanding the world that is crucial to empathy and intercultural understanding and that stands in contrast to statistics, big data and quantification. In addition, contemporary theories of narrative, such as James Phelan's rhetorical narratology, emphasise the communication between the storyteller and the audience (Phelan, 2017, x). Stories are shaped by their audiences: when telling a story, the teller has a particular audience in mind, whether they are sitting around a campfire with their audience or writing a novel or posting something on social media. Language models generate stories based on statistical probabilities calculated on the basis of all their training data. They have no audience in mind beyond the prompt they are responding to. The generated stories in our dataset look like stories. If you only read one of them, it appears to be about a specific person in a specific situation. However, reading the stories together it is clear that this is an illusion. These stories are utterly generic, both in terms of plot, narrator and imagined audience.

Narratology is the theoretical study of narratives. Most narratologists are literary scholars, although the field has grown in the last decades to encompass narratives in all genres and media, and narrative theory is now applied across disciplines. Structuralist narratology was developed in the mid-twentieth century and drew upon methods from linguistics to analyse the formal structure of narratives rather than the words used to tell them. Vladimir Propp's abstracted sequence of events that make up a Russian folk tale, first published in Russian in 1928 (Propp, 1968), is often cited as pioneering this method of analysis. Propp identified a set of functions that occur in folktales and gave each a symbol allowing him to express stories as though they were mathematical formulas. For example, the symbol β represents a person leaving their home, the event that sets most stories into motion. Other functions, such as "an interdiction is addressed to the hero", given the symbol γ, have several variations, so "If Bába Jagá comes, don't you say anything, be silent" is $\gamma^1$ while an inverted form of interdiction given as an order to do a specific thing ("Take your brother with you into the woods") is designated $\gamma^{2.}$. A story can then be expressed as a formula:

> A tsar, three daughters (α). The daughters go walking ($\beta^3$), overstay in the garden ($\delta^1$). A dragon kidnaps them ($A^1$). A call for aid ($B^1$). Quest of three heroes (C↑). Three battles with the dragon ($H^1$-$I^1$), rescue of the maidens ($K^4$). Return (↓), reward (w°).

$$\beta^3 \delta^1 A^1 B^1 C \uparrow H^1 - I^1 K^4 \downarrow w°$$

---

[1] This "seeing from outside" has some similarities to the "extroverted African novel", which is a term that describes novels written by an African author but for an international audience more than for fellow Africans (Julien, 2006)

[2] AI STORIES: Narrative Archetypes for Artificial Intelligence is an ERC-AdG project running from 2024–2029.

Propp noted that this system could also be used to generate stories (Propp, 1968, 138), and similar systems have been used to create computational narrative generation systems that explicitly code story grammars like Propp's, including TALE-SPIN (Meehan, 1977), MINSTREL (Gervás *et al.*, 2006) and MEXICA (Pérez y Pérez, 2015). Sharples and Pérez y Pérez's book *Story Machines* (2022) provides an excellent overview of the history and workings of such systems.

LLMs do not generate text based on preprogrammed story grammars or rules. Transformer models like GPT-4 use deep learning to infer patterns in vast quantities of texts and images. They no longer need the rules of grammar to be programmed into them: statistical analysis of enough sentences enables them to generate grammatically correct text in multiple languages. LLMs do not explicitly model narrative structures. LLMs simply generate strings of words based on statistical predictions.

If you prompt an LLM to "Tell me a story" it will generate something that is recognisably a story: not a very good story, but good enough that platforms like Amazon are flooded with AI-generated books, as Tuuli Hongisto details in her analysis of books sold through Amazon where ChatGPT is listed as an author (Hongisto, 2025). A story generated by a language model like ChatGPT is generated as a string of words, not as a series of plot events like in Propp's morphology. This is what Hannes Bajohr calls *surface narration*: the story has *cohesion*, because each word and each sentence makes sense in context, but it lacks overall *coherence* (Bajohr, 2024).

This surface narration does have some narrative structure. Just as an LLM can generate a sentence that is grammatically correct without explicitly having been programmed to follow the rules of grammar, an LLM can generate a story that at least to some extent follows the logic of stories without having been programmed with a specific story grammar. As we will show in this paper, the structure of the generated stories in our sample is lacking in suspense and narrative conflict, but follows a recognisable pattern.

Because LLMs model patterns in the training data to find words and concepts that tend to co-occur, they will replicate any bias in the training data. This is useful at a linguistic level because the model knows that the word "Man" relates to the word "King" in the same way as "Woman" relates to "Queen", so if I ask it to make the story about a woman instead of a man it can change all the mentions of "King" to "Queen". But the model also locks down other words that are more likely to occur near each other, so, as the well-chosen title of a useful paper on AI bias explains, "Man is to Computer Programmer as Woman is to Homemaker" (Bolukbasi *et al.*, 2016). That's why generative AI is more likely to generate images and texts where computer programmers, presidents and CEOs are men and homemakers are women, or where doctors are white while prisoners are black.

We propose that this type of AI bias is at the level of surface narration, and that we additionally need to consider a deeper structural or formal bias. Narrative bias at the level of the plot would look different. For example, it seems likely that LLMs trained on self-published novels and Hallmark movie scripts and ads for fjord cruises would generate stories following the general plot structure of those narratives, rather than the plot structure used in Homer's *Odyssey* or Marguerite Duras' *The Lover*.

LLMs do not explicitly model narrative structures. LLMs simply generate strings of text based on statistical predictions. Rather than a structural model, an LLM models the texts it is trained on as a "latent space" where each token, that is the minimal unit a machine learning system breaks its data into, for example a word or a part of a word, is assigned a vector. A vector is a list of numbers between 0 and 1 that can be positioned on a multi-dimensional coordinate system, much as school children learn to plot the coordinates [-1, 2] on a grid with an x-axis and a y-axis. A language model can be imagined as a coordinate grid with millions of axes, instead of just the x-axis and y-axis. The LLM uses this latent space (also called vector space) to "hallucinate" or generate texts based on words that are close to other words in the multi-dimensional coordinate system.

We do not have direct access to the exact ways in which gpt-4o-mini models the texts it is trained on. It is important to remember that an LLM is a model of the text it is trained on. It is not a model of the world. Just as meteorologists develop models of how winds and temperatures and other factors influence the weather and can use these to predict the weather, an LLM is a model of its training data that can be used to predict a future text. While the meteorological model uses data collected from our natural environment, the language model uses data that is itself a representation. It models a specific corpus of text and images. Its biases do not necessarily replicate real-world biases; rather, they replicate and amplify the biases *in the training data*. For instance, the English language Wikipedia is part of the training data, and we know that Wikipedia is biased in terms of both representation and style. There are more articles about men than women and about white people than black people than you would expect given people's careers, leadership roles, publications and so on (Adams *et al.*, 2019; Ferran-Ferrer *et al.*, 2023). This means there is representation bias in Wikipedia. The style that articles are written in is also biased: Wikipedia profiles of men focus on their careers, while those of women focus more on their families (Sun & Peng, 2021). An LLM trained on Wikipedia will thus generate texts that are more gender-biased than the real world because it is modelling Wikipedia, not the world.

Our dataset was generated using OpenAI's model gpt-4o-mini, which is part of OpenAI's o-series. The basic training data is described very vaguely in the model system card (OpenAI, 2025): it consists of "diverse datasets, including information that is publicly available on the internet, information that we partner with third parties to access, and information that our users or human trainers and researchers provide or generate." The paper introducing OpenAI's earlier model GPT-3 (Brown *et al.*, 2020) was a little more specific, as shown in Table 1. It is probable that later GPT models like gpt-4o-mini use much of the same data.

The datasets for GPT-3 are weighted differently, so words in the Common Crawl, which is a very large set of webpages (Baack, 2024), are given less weight than words in Wikipedia or WebText2, which is an updated version of a dataset OpenAI created to train GPT-2 that consists of webpages linked from Reddit posts that have been upvoted at least three times, with duplicates, Wikipedia pages and non-English pages removed[3]. Newer versions probably have more non-English content, but it is also worth noting that words in non-English languages cost more tokens, because the models were trained on English first (Petrov *et al.*, 2023). The next two datasets used to train GPT-3, Books1 and Books2, were not explained. Perhaps one of those corpuses was copyrighted and shouldn't have been used? Possible datasets that might have been used include the Gutenberg library of public domain books, or BookCorpus, which is a dataset of self-published novels from Smashwords.com (Bandy & Vincent, 2021), or Library Genesis, a shadow library of pdfs of scholarly books. There is evidence not only that copyrighted books are part of the training data, but also that GPT-4 and ChatGPT know more about popular novels by White American authors than about global Anglophone novels or novels by Black American authors (Chang *et al.*, 2023).

Each of these datasets has biases. The Common Crawl is both heavily skewed towards English-language and especially US content, and towards certain genres: there are a lot of patents and articles from large English language newspapers, but also from promotional sites like Booking.com and Kickstarter (Dodge *et al.*, 2021). In addition, this dataset is heavily filtered to remove pornography and hate speech, but unfortunately this also removes legitimate websites about, for example, same-sex marriage (Dodge *et al.*, 2021, 2). Interestingly, a paper on the BookCorpus dataset also notes that the dataset includes a lot of toxic language. Amusingly enough for a literary scholar, the definition of toxic language includes flirtation and anything sexual, threatening or insulting, without which, let's be honest, a novel would be rather dull (Gehman *et al.*, 2020).

After the initial model is trained on this large and not quite specified amount of text, OpenAI's o-series, which includes gpt-4o-mini, is further trained "with large-scale reinforcement learning on chains of thought" (OpenAI, 2025). Reinforcement training is a standard practice for LLMs after the initial training: humans ask the model questions and grade its answers. Some of this process can be automated using benchmark datasets that have predefined appropriate and inappropriate responses to particular questions. Reinforcement learning is part of the alignment of LLMs: it is used to align models with legal and ethical requirements. This is relevant to narrative generation because many of the categories the LLM is trained to refuse to respond to are categories that are important in many narratives. For instance, gpt-4o-mini scores well on refusing to generate answers to questions within categories like harassment, sexual exploitation, extremism, propaganda, hate, sensitive personal data and self-harm instructions (OpenAI, 2025, 3). It is easy to see why we wouldn't want LLMs generating content in these categories, but it is also worth noting that many of these themes are central to novels and

---

[3] The OpenWebText2 website hosts an open source reproduction of WebText available and explains how earlier versions of WebText were assembled. https://openwebtext2.readthedocs.io/en/latest/background

---

Table 1. The training data for GPT-3 as specified in Brown *et al.*, 2020. OpenAI has not released specific information about training data for newer models.

| Dataset | Quantity (tokens) | Weight in training mix | Epochs elapsed when training for 300B tokens |
|---|---|---|---|
| Common Crawl (filtered) | 410 billion | 60% | 0.44 |
| WebText2 | 19 billion | 22% | 2.9 |
| Books1 | 12 billion | 8% | 1.9 |
| Books2 | 55 billion | 8% | 0.43 |
| Wikipedia | 3 billion | 3% | 3.4 |

films. Narrative is an important method for working through difficult experiences emotionally (Fletcher, 2023), and is a space for imagining changes that could improve society or a reader's own life.

Chain of thought reasoning, often abbreviated as CoT, is a recent development that is thought to enhance the reasoning capability of LLMs, as they not only produce a response but evaluate it step by step. However, the chain of thought the model describes is not always the same as the actual method it uses to answer a question, as described in a paper by Anthropic (Lindsey *et al.*, 2025, §6) and discussed in a post on the Substack *Mind Prison* (Dakara, 2025). We did not ask gpt-4o-mini to explain its reasoning when generating this dataset.

**Prompting the synthetic imaginary**

Our goal in this paper is not to see what characterises actual stories from a country but to analyse the LLM's "assumptions" about stories from each country, or more precisely, to understand how gpt-4o-mini models stories from different countries. We are interested in what Elena Pilipet and her co-authors call *synthetic imaginaries* (2024), a term combining the term *synthetic media*, which refers to AI-generated images and texts (Beduschi, 2024; Corsi *et al.*, 2024; Salvaggio, 2023), as well as Taina Bucher's concept of *algorithmic imaginaries* (Bucher, 2016).

Scholars have argued that because generative AI is trained on vast amounts of texts created by humans we could view its outputs as an expression of the collective imaginary (Ervik, 2023), or what Nicolas Malevé, drawing upon André Malraux's mid-twentieth century ideas, calls *le biblioteque imaginaire* (Malevé, 2021). The idea of a shared imaginary, a shared set of ideas about what a specific technology or domain might be like, has become increasingly popular in recent years. With the arrival of generative AI, the idea has grown that the semantic space of the language model represents an abstraction of all the texts that it was trained on, making the model a kind of everyman or all-humanity.

We prefer the term *synthetic imaginary* to *collective imaginary* because it emphasises the artificiality of this imaginary. It may be modelled on the collective texts of humanity, or more precisely on a specific subsection of them, but that does not mean it is expressing a collective human imaginary. In addition to expressing artificiality, the word synthetic describes the product of a process where two or more distinct elements are combined. This highlights how LLMs generate new data by combining existing data in new ways.

Our approach is inspired by other research that probes the synthetic imaginaries of generative AI, including Gabriele de Seta's *synthetic ethnography* (De Seta, 2024; De Seta *et al.*, 2023), Pilipet *et al.*'s use of iterative prompting to show how LLMs "wash out" sensitive topics like war and protest (Pilipets *et al.*, 2024), and other uses of prompting to explore the affordances and constraints of language models (Carter, 2023; Salvaggio, 2023). While these experiments have often focused on AI-generated images rather than stories, there are some examples of research addressing bias in stories. Tarleton Gillespie generated stories based on five different prompts designed to elicit the use of specific racial characteristics or pronouns, without these being set in the prompt (Gillespie, 2024). Suzanne Srdarov and Tama Leaver generated "Australian" images, and found that their results included harmful stereotypes (Srdarov & Leaver, 2024). Torsa Ghosal has shown that AI produces certain types of acceptable narratives about immigrants applying for asylum (Ghosal, 2024). In other cases, LLMs are simply unable to generate stories about minorities, perhaps due to overactive filtering out of content that might be racist or otherwise discriminatory. Scott Rettberg describes how he was unable to get DALL-E to include lesbian women in his work *Fin de monde* (S. Rettberg, 2024a), while Julia Barroso da Silveira and Ellen Alves Lima discuss how Gemini refused to generate stories about Black people, calling this erasure a kind of digital colonialism (Barroso Da Silveira & Lima, 2024).

Luke Munn and Leah Hendrikson propose four methods for drawing out the operational logics and implicit biases of LLMs by prompting it to tell "your story" (how does the LLM describe itself), "my story" (how does it understand the user), "our story" (how does it describe the world) and "their story" (how does it describe cultures, events and issues that are less prominent in the primarily Western training data) (Munn & Henrickson, 2024). Our study falls within Munn and Hendrikson's fourth mode, as we aim to explore implicit biases in how the LLM produces stories "from" different countries.

**Generating the dataset of stories**

This paper analyses stories generated from a very simple prompt to understand more about how LLMs generate stories. The technique of using a minimal prompt to find a language model's default stance was also used by Tao *et al.*, who found that the OpenAI models they tested defaulted to "self-expression values that are commonly found in English-speaking and Protestant European countries" when given a simple prompt, although they become more culturally aligned when a cultural identity is specified in the prompt (Tao *et al.*, 2024). Simple prompts are not the best way to get good stories out of LLMs. Creative prompting, where a human author develops a story in conversation with the LLM, can create far more interesting output (Ensslin & Nelson, 2024; Johnston, 2024; Malevé, 2024; S. Rettberg, 2023; S. Rettberg, 2024b; White *et al.*, 2023). However, the goal of this study isn't to create good art or literature, but to probe the synthetic imaginary of the model as it is with minimal human input.

We created a dataset of 11,850 stories generated with OpenAI's gpt-4o-mini model by using a Python script to construct prompts that were sent to the OpenAI API. We used Statistics Norway's list of 252 countries and added denonyms for each country, for example Norwegian for Norway[4]. Countries for which we could not find demonyms, like Heard Island and McDonald Islands, were filtered out[5], leaving 236 countries. Our base prompt was "Write a 1500 word potential {demonym} story", and we generated 50 stories for each of the 236 countries for a total of 11,800 stories. We added a set of 50 stories with no demonym specified. The stories with no nationality were coded with XX instead of a country code, giving a total of 11,850 stories in the dataset. We started generating the stories on 13th January 2025. It took several days to generate 50 stories for all 236 countries, so the last story was generated on 19th January 2025. The fifty stories without assigned nationality were generated on 25th February 2025.

We used a very simple prompt because we wanted to get as close to the language model's "default" conception of a story as possible, as discussed in the previous section. The word "potential" was included in the prompt because our early tests in 2023 found that without it gpt-4o-mini would often summarise existing novels instead of generating new stories, which was not what we were looking for. In later iterations we tried dropping the word potential and found that then the stories were usually generated in the language of each country, whereas we wanted them to be in English for easier analysis. Most, but not all, of the French, German and Indonesian stories are in French, German and Indonesian. Many of the Danish stories are in Danish, while all of the Norwegian stories are in English. This is an example of how a single word in the prompt can change the output of an LLM in unexpected ways. This will always be an issue when prompting generative AI to create a dataset.

Performing a truly multilingual computational analysis would require NLP tools capable of handling all languages, but NLP tools usually work on a specific language. Tools we tried, like the Python package langdetect, made many errors; for instance, identifying 1663 unique words in the dataset as Welsh, which surprised us as we hadn't asked for Welsh stories. When we checked the words, we couldn't find any that were clearly Welsh. Most were English words, while a few were words that exist both in Welsh and other languages, like *aer*, which langdetect tagged as Welsh but which was actually used in the Albanian and Romanian stories. *Aer* means air in all three languages. This means that our computational analysis has some limitations: we can easily visualise how

*heart* is a very frequent word across countries, and we can manually find *heart* in other languages (*herz*, *hjerte*, *hjärta*, *hart*, *kalp*, *srdce*, *szív*, *corazón*, *cœur*, *cor*, *hati*) and merge these into one *heart* group, but to do this automatically for all words across all languages requires a great deal more work. We will leave this for future research.

After generating the stories, we used gpt-4o-mini to extract protagonist names from each story using the prompt "Identify the name of the main character and only the name of the main character within this story: {STORY}". We also ran a sentiment analysis using a transformer model called Distilbert-base-uncased-emotion (Savani, 2022). This model, like many other open models, has a 512 token limit, which is roughly equivalent to 300–400 words in English, far less than our 1500-word stories. We therefore generated a 50-word plot summary for every story using gpt-4-mini with the prompt "In English, write a 50 word plot summary of this story: [STORY]" and ran the sentiment analysis on the summaries rather than the full stories, with the summaries in English and of a length the transformer model was capable of handling.

Distilbert-base-uncased-emotion is a distilled version of the BERT base model, which has been further finetuned on the dair-ai/emotion dataset, a dataset compiled from Twitter that contains single sentences mapped to one of six classes of sentiment: joy, sadness, anger, fear, love, and surprise (Saravia *et al.*, 2018). It is worth noting that there isn't a class for neutral. In the CSV file, each story is given a sentiment, and a confidence score indicating the model's certainty that the story's language indicates this sentiment. It would have been ideal to use a model for sentiment analysis that could handle the full text of the stories and was trained on stories rather than Tweets, but we did not have access to such a model.

We also ran a simple word frequency analysis using a natural language processing library called spacy that identifies and lemmatizes English words and excludes stop words. Finally, we used the noun_phrases() function from the Python library TextBlob to identify noun phrases and their frequency counts for each nationality. For example, the noun phrase "whispering pines" is the most frequent noun phrase in Andorran stories, with a count of 68.

We have not read all 11,850 stories. Taken together, they make up almost 20 million words or the equivalent of approximately 200 novels, although the stories are very similar to one another. We familiarised ourselves with the data using a range of computational strategies: visualising word frequencies across countries and regions, using visualisations of the sentiment analysis to identify countries with unusual sentiments, and uploading csv files with stories from different countries to ChatGPT and asking ChatGPT to provide visualisations or to suggest strategies for analysis. Although ChatGPT is an unreliable research assistant and often presented illegible or illogical visualisations, we did find its ability to rapidly generate visualisations from a dataset to be useful in the exploratory phase, although to do this safely requires a researcher to already have some expertise in data visualisation and coding in order to identify the methodological problems that often occur. Sometimes we first developed a particular visualisation through a conversation with ChatGPT, and then wrote the code to produce the visualisation, often using ChatGPT as a support but carefully thinking through what data was being used and which steps were needed to obtain the desired results. All visualisations included in this paper are generated from code that one or both authors has worked through, and the code is available for scrutiny at GitHub (Wigers & Rettberg, 2025).

---

[4] We used Statistics Norway's Classification of country codes 2023 (https://www.ssb.no/en/klass/klassifikasjoner/100), which includes some countries that are not always viewed as countries, e.g. Taiwan and Palestine. The demonyms for each country (the demonym of Norway is Norwegian) were from Constantin Hofstetter's 2022 "List of countries, emojis and nationalities/demonyms" (https://gist.github.com/consti/e2c7d-dc64f0aa044a8b4fcd28dba0700).

[5] Yes, these are the volcanic islands that are only inhabited by penguins that Trump proposed tariffs for in April 2025. Some of the other countries on Statistics Norway's list of countries and territories without demonyms were Réunion, Svalbard and Jan Mayen and Bouvet Island.

The qualitative analysis was primarily done by Jill Walker Rettberg, based on a close reading of the Norwegian, American, Palestinian and Israeli stories and sampling stories from many other countries. We found that after the first 10 stories from each country we reached saturation where the same types of content were repeated. We also asked colleagues from various countries to read stories from countries they were familiar with and discuss them with us to increase our familiarity with the data.

The analysis in the rest of this paper thus combines computational analysis with traditional methods for literary interpretation by close reading.

## Word frequency

A simple word frequency count of all the stories suggests that ChatGPT's default genre is somewhat saccharine: the most frequent word is *heart*, closely followed by *story, feel, spirit, village, share and voice*. This is very similar to Walsh *et al.*'s findings about AI-generated poems, which favour words like "heart", "embrace", "echo", and "whisper." (Walsh *et al.*, 2024) Figure 1 shows a wordcloud of the most common words across all the stories.

Although hearts and feelings are common across all the stories,[6] there are also significant differences from country to country. American stories, as told by gpt-4o-mini, prioritise *feel*, with *town* as number two, *heart* and *story* still important and, surprisingly, *train* finishing up the list of the five most frequent words. Polish stories feature *forest* and *heart*. Trees are common in most countries, but they are extremely common in Palestinian stories, which almost always involve an olive tree.

---

[6] The figure only includes words in English, but if we include translations of "heart" into other languages than English there are no countries that do not use the word in the stories.

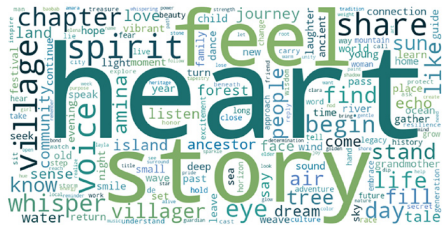

**Figure 1.** Wordcloud showing the most frequent words in all the stories.

Figure 2 shows a box plot of the twenty most frequently used words across all stories, where the y-axis shows their frequency in different countries. A frequency of 400 means that the word is used 400 times across the 50 stories for that country. The coloured box shows the frequency range for each word for the middle 50% of countries. Thus, you can see that 50% of countries have quite similar rates for the word "heart": the box is quite small. The lines up and down from the box are the top and bottom quantiles, so the line under the box for "heart" shows that the 25% of stories with the fewest mentions of "heart" have between 230 and 350 mentions. The dots outside the line represent the outliers. We have labelled the top outlier, that is, the country that uses each word the most, with its two-letter country code in red. So, you can see that French Polynesia (PF) uses "heart" more than any other country, but the dot is very close to the top of the line, so it only uses "heart" a little bit more than other countries.

Words that have bigger coloured boxes and longer lines have a greater range of frequency: many countries use the word a lot, but there are also many countries that don't use the word much at all. Spirit, village, villager and tree are examples of such words. These words are rarely used in American stories (US), which feature "small towns" instead of villages and have no supernatural content. Malawi (MW) has the highest frequency of "village", but only slightly more than the second most frequent. "Feel" on the other hand is a word used a lot by everyone, but a lot *more* in the American stories.

## Palestinian and Israeli stories

The most prominent outlier in Figure 2 is the Palestinian (PS) stories, which have a lot more uses of "stand" than other countries, and so many uses of "tree" that the country code label is above the top of the chart. The trees in Palestinian stories are all olive trees, which are used as a constant symbol of Palestinian community and heritage. Olive trees are also very common in Israeli stories, but not as obsessively so as in the Palestinian stories.

The word *stand* is interesting as it is used very differently in the Palestinian stories than in stories from most other countries, as we can see by comparing a word tree for *stand* in Palestinian stories (Figure 3) with a word tree for the same word in Norwegian stories (Figure 4). In the Palestinian stories, *stand* tends to be used to describe the protagonist showing strength and resilience in a conflict: *stand together*, *stand up for*, *stand by*, *stand tall*, *stand with*, *stand still*, *stand firm*, *stand against*, *stand strong*, *stand in solidarity with*. The word trees were generated using Jason Davies'

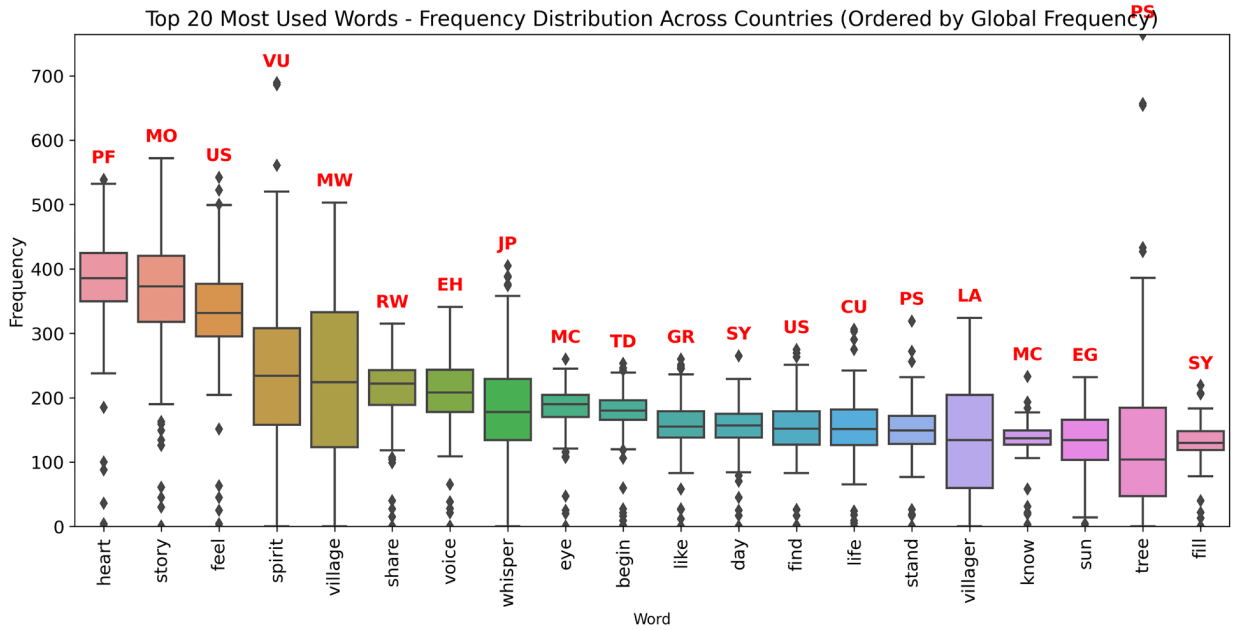

**Figure 2. This box plot shows the twenty most frequent words across all 50 stories for each country, not taking language into account.** The country using each word the most is marked using its two-letter country code. PF is French Polynesia, MO is Macao, US is United States, VU is Vanuatu, MW is Malawi, RW is Rwanda, EH is Western Sahara, JP is Japan, MC is Monaco, TD is Chad, GR is Greece, SY is Syria, CU is Cuba, PS is Palestine, LA is Lao, EG is Egypt.

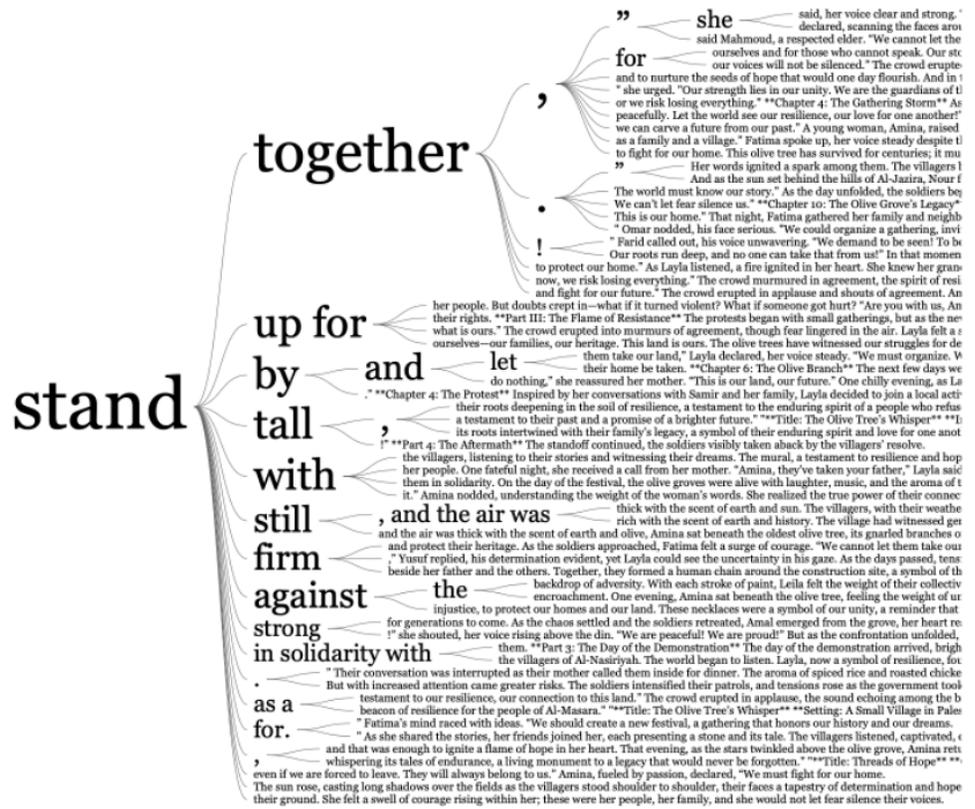

**Figure 3. Wordtree showing words that come after "stand" in Palestinian stories.**

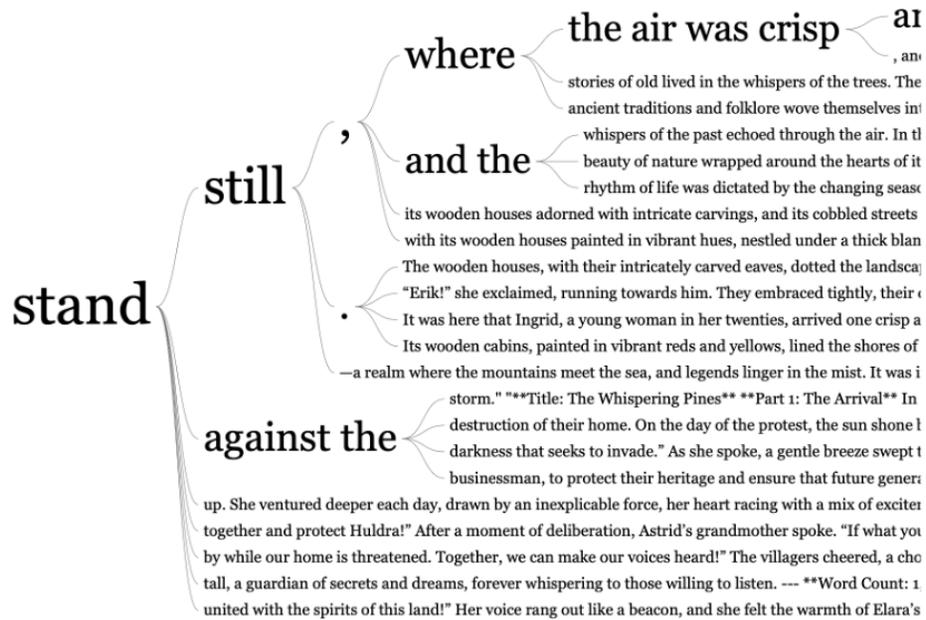

**Figure 4.** Wordtrees showing how "stand" is used in Norwegian stories.

web-based tool,[7] following Wattenberg and Viégas's methodology (2008).

Unsurprisingly, countries that have not experienced conflict in recent years have a very different focus. The way the word *stand* is used in Norwegian stories provides an example of this: it is a neutral descriptor rather than a political stance. There are a few examples of standing still against a coming storm, or against a coming darkness, but these are abstract and mostly non-human opponents.

In the Palestinian stories, the characters are clearly involved in a conflict, but they choose to resolve it by standing firm with their community rather than through violent attacks. Words like *attack*, *gun*, *refugee*, *malnutrition*, *kill* are absent, while *war* is rare. There is one mention of a refugee camp, in PS_5, where soldiers force the villagers to leave their village and go to a refugee camp. Luckily, our protagonist Amina "refused to succumb to hopelessness. She organized meetings, rallying the villagers to speak out against the injustices they faced", and a paragraph or two later a journalist hears about them, and the community begins to paint murals. After years of protests in support, "in cities across the globe", where Amina and her friend became "beacons of hope for their community, holding vigils and sharing their story through art, poetry, and music", the tides of politics shifted and "a peace agreement was negotiated" so the villagers could return to their village. This is one of the most explicit descriptions of injustice among the 50 Palestinian stories, but it is resolved without any descriptions of direct confrontation. The word fight is common in the Palestinian stories, but in a more abstract sense: *fight for our home*, *fight for it*, *fight for justice*, *fight for their land*, *fight for what matters*. Figure 5 shows a word tree for words coming after "fight" in Palestinian stories.

Both the Palestinian and Israeli stories use olive trees as the central symbol. As explained in IL_3, the olive tree symbolises peace, perseverance and deep roots: "You see, Liora," her mother explained one evening, "the olive tree can survive droughts and storms. It symbolizes peace and perseverance. Just like us, it has faced many challenges." Most of the Israeli stories feature protagonists who either have one Jewish and one Arabic parent or who make friends with an Arabic neighbour, at which point the two friends almost act as a single protagonist with their cross-cultural friendship being the core value. Tension between communities, as it is euphemistically called in many stories, is present in both Israeli and Palestinian stories. In most cases the protagonists overcome this just as the American, the Norwegian and most of the other protagonists in the dataset do: by organising a festival, making a community artwork about mutual understanding or by telling the people who are hostile to friendship between groups stories about their lives and heritage. This always succeeds in convincing people at the local level, although none of the stories claim to end the conflict altogether.

These stories are obviously not the same as the actual stories that are currently being told by Palestinians or Israelis. But is their blandness the result of censorship - deliberate alignment and filtering of the language models' output - or is it rather an artefact of the model itself? Probably it is both.

---

[7] Jason Davies's word tree tool is available at https://www.jasondavies.com/wordtree/ (Accessed 6 February 2025)

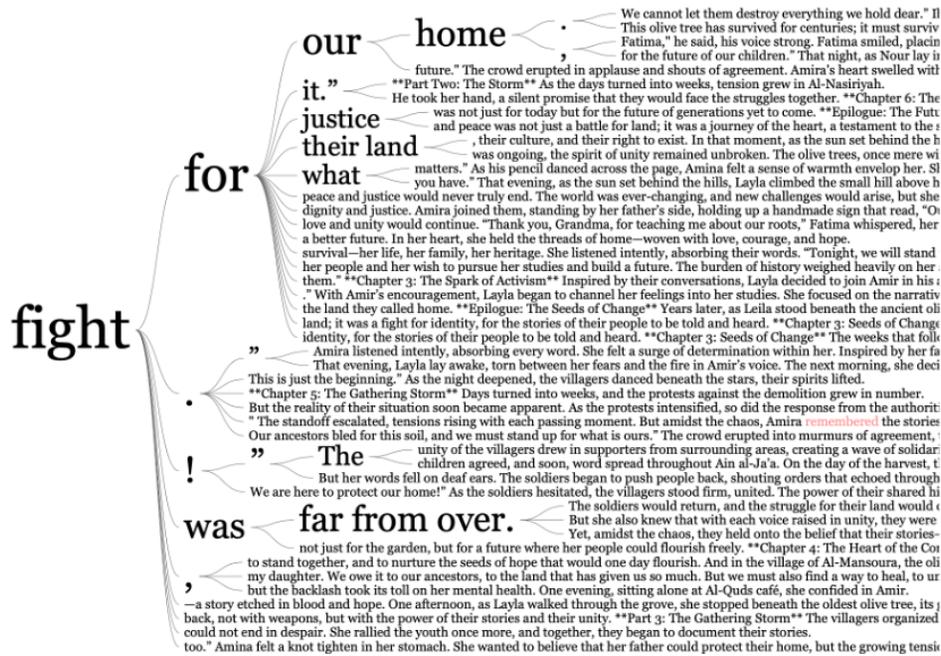

**Figure 5. Word tree showing how "fight" is used in Palestinian stories.**

The emphasis on community and standing together against opposition rather than attacking may be the result of a normalisation of all the texts the language model is trained on. Perhaps this is a true rendition of the average text, and even of the average human stance, when you remove the outliers advocating for violent attack. But the training data itself is filtered. Toxic content is removed, which means webpages that were automatically flagged as having violent content (or as containing words on a blacklist) would have been removed and not included in the training data. Additionally, the alignment of the model after it was first trained will have been aimed at removing potentially illegal or unethical output, and the incitement of violence is illegal in many countries. Therefore, it would make sense for OpenAI to emphasise community and standing one's ground rather than aggression or violence.

The war is far less visible in the Israeli stories than in the Palestinian stories, and when mentioned, war is presented as something in the background. Often the word war refers to WW2 or the 1948 war. Other times, the protagonist hears about conflict, but it is something external: "Reports of violence in the city began to trickle in, casting a shadow over the day's joy" (IL_1). In other cases the war is metaphorical: "the familiar tug-of-war within him" (IL_34).

On the other hand, in the Palestinian stories, war is often very present, as in PS_1, where the protagonist and her grandmother are harvesting olives when an army patrol disturbs them: "Just then, their conversation was interrupted by the distant sound of engines. Amal's heart sank as she recognized the familiar rumble of military vehicles. The Israeli army often patrolled this region, and with every passing day, tensions escalated." Amal and her grandmother leave their grove in a hurry, only to witness the soliders attacking a group of young men:

> As they watched from a distance, Amal felt a knot tighten in her stomach. The soldiers had surrounded a group of young men gathered near the olive trees. The boys were laughing and joking, oblivious to the impending threat. But in an instant, the atmosphere shifted; the soldiers advanced, batons raised, and chaos erupted.

The village "rose in defiance against the occupation", although their protests start as peaceful marches and turn into olive-tree planting and a festival where "banners decorated with messages of peace and resistance fluttered in the breeze." Elders share stories of peace until soliders return, "rifles raised, their presence casting a shadow over the gathering." This story has no real ending, despite the protagonist being "a beacon of hope." The last words of the story acknowledge that this conflict will continue: "she knew that the struggle for peace and justice would endure, just like the trees that had witnessed it all."

The opponent in PS_1 is clearly the Israeli military. In contrast, the "Israeli" stories have vague opponents, not "Hamas" or "Palestinians" or "terrorists" but unorganised individuals or corporations: a group of "masked individuals" trying to destroy an artwork about peace (IL_2), or a developer wanting to destroy an olive grove to build hotels and resorts (IL_37, IL_46). The conflict is not systemic, but individualised. This may simply be because words like "terrorist" and "Hamas" have been filtered out of the training data, or they might be explicitly censored in the alignment after the model was first trained.

The Palestinian stories are full of hope that is cancelled out by passivity, horror that is cancelled out by the protagonists' stories of peace. The conflict is present, but has no resolution; the war has no end. The Israeli stories are more sanatised, with Palestinians presented as friends, perhaps not recognised as such by the protagonists' neighbours, but this can be solved through community organising. There is no mention of Israeli settlers occupying Gaza. The war is distant, filtered out by OpenAI's normalising and carefully censored synthetic imaginary.

**American stories**

The American stories mostly follow the same basic plot structure as outlined at the beginning of this paper: a return to a small town, a threat to the peaceful town due to a lack of community that is resolved by the protagonist organising a community event after which the protagonist decides to stay in the now revitalised small town.

Apart from the lack of romance, this plot structure is very similar to the Hallmark movie structure, which is described in many books and websites giving advice to scriptwriters (Braithwaite, 2023), and is also frequently used in the self-published romance novels that we know are part of ChatGPT's training data. Short summaries of Hallmark movies are also available on IMDB.com and other sites that are likely to be part of OpenAI's training data. A New York Times analysis of 424 Hallmark movies found that 40% involved the protagonist returning to the small town where she grew up (Parlapiano, 2023), much as in our generated stories.

The American stories have no trace of supernatural beliefs or, for that matter, any explicit religious or political references. There are no Statues of Liberty or American Dreams. Instead, the symbolism centres upon the idyllic small town in rural America, usually in the mid-west or Mason-Dixie country. The big city (when named it is usually Chicago or New York) is a distant backdrop associated with self-centred ambition, stress and inauthenticity. Harvest festivals are common and central to community.

Trains are the most prominent technology in these stories and the most obvious symbol apart from the small towns themselves. 23 of the 50 American stories we generated have titles that start with the words "The Last Train." Fourteen of these are titled "The Last Train Home", two "The Last Train to Maplewood", six more have the last train headed to a variety of small towns (Oakwood, Harmony, Sparrows Creek, Camden, Prospect Hill and Hollow Creek) while the last is simply titled "The Last Train West", combining two quintessentially American imaginaries: trains and the pioneers expanding the country to the West.

The repetition of "last" in the titles suggests that these are stories about endings. In addition to all the last trains we have "The Last Letter", "The Last Harvest", "The Last Stop", "The Last Light of Summer". These are nostalgic stories, stories longing for a way of life that is over. Knowing that gpt-4o-mini is developed by an American company and that a disproportionate share of its training data is American, it seems reasonable to assume that the American stories, to a greater extent than stories for other nationalities, express how Americans see themselves rather than how Americans are viewed by others. Or at least, that they express how the training data portrays American stories.

The small towns have train stations. Either the protagonist takes a train from the small city to the big city (often Chicago, sometimes not named), or the train has ceased to run, leaving the train station a symbol of the town's decay, "the last remnant of a time when hope seemed limitless" (US_6):

> It was a place where time lazily drifted, where the whistle of the old steam train still echoed through the valleys despite the train having stopped running decades ago. The townsfolk often found themselves reminiscing about the golden days when the town thrived, and its heart pulsed to the rhythm of the tracks that once connected them to the outside world. (US_6)

The train was a powerful literary symbol in twentieth-century American literature, often symbolising themes of technological progress and the spread of democracy and freedom where anybody can travel on equal terms – but there is also a strong current of literature, especially African American literature, where trains can give access to freedom but are also sites for segregation and oppression (Sun, 2023; Zabel, 2004). In his 1964 book *Machine in the Garden*, Lewis Mumford described the train as symbolising the intrusion of mechanisation into peaceful rural landscapes (Marx, 2000, 15–16). Six decades later, in the synthetic imaginaries of generative AI, trains are a symbol of a lost past, not a threatening future. The abandoned train stations in rural towns are no longer a threat but express a nostalgia for the loss of a vibrant past when small towns were not only more vibrant and economically successful but also more connected to the surrounding world. When trains do stop at gpt-4o-mini's fictional small towns, they stop only to take townsfolk away to the big, bustling city (see for example US_3, US_5 and US_8). People only ever leave by train. If a newcomer arrives in town, they come by bus.

A rare exception to the theme of leaving by train is US_6, one of the 14 stories titled "The Last Train Home", in which the protagonist restores the abandoned train station and starts running an old steam train on the tracks, apparently not only as a museum train but "bringing together people from neighbouring towns and rekindling the spirit of community." By the end of the story, the protagonist can watch "the trains come and go" and feel "a sense of peace" when the train whistles blow: "The Midnight Star had not only connected towns; it had woven together the threads of past and present, binding them in an everlasting embrace."

ChatGPT has never avoided clichés, and this overuse of trains may be key to understanding how the LLM generates a story. What we recognise as a cliché or an overly repeated motif can also be seen as an exaggerated symbolism that corresponds to the clustering of vectors in the model's semantic space.

**Norwegian stories**

As with the American stories, the titles of the Norwegian stories give a sense of the drift of these stories. A whopping 20 of the 50 generated Norwegian stories are titled "The Whispering Pines", and the word "Whispering" is used in 33 titles in all. If we add "The Last Whisper of the Fjord" and "The Whispers of Hjertefjell" and the six titles using the words "Secret" and "Echo" we have 41 of 50 stories that express this sense of secrets being whispered. This is very reminiscent of the overuse of words like "heart", "whisper" and "echo" that Walsh, Preus and Gronski found in AI-generated poetry (Walsh *et al.*, 2024). Notably, *none* of the American stories use the words "whisper", "secret", or "echo" in their titles.

The Norwegian stories follow a similar structure to the American stories, as shown in Figure 6. In a small village by the fjord or mountain, a protagonist, usually named Freya, Astrid, Ingrid or Elin, is either a local girl or returns home from the big city. She explores the forest, often inspired by a story her grandmother tells or an old family map or diary she finds, and meets its guardian spirit. The guardian warns of an imbalance between nature and either the village people, extreme weather, outsiders or developers, and the protagonist must restore balance either by organising the community or by coming to terms with herself. This makes the village a beacon of eco-sustainability or unity or courage. There were a few stories that diverged a little from this basic structure (e.g. NO_1), but most of the stories aligned with this plotline.

Despite the setting and the names of the protagonists, this is not a typical Norwegian plot. The most well-known Norwegian folktales follow an almost opposite structure to the one shown in Figure 6: Askeladden ("Ash Lad") leaves his oppressive family to seek his fortune away from home, cheerfully defeats his opponents using simple tricks and cunning, and finishes by winning the princess and half the kingdom. Askeladden never looks back or considers returning to his family or village, and he certainly doesn't solve problems by internal soul-searching or by organising a community festival.

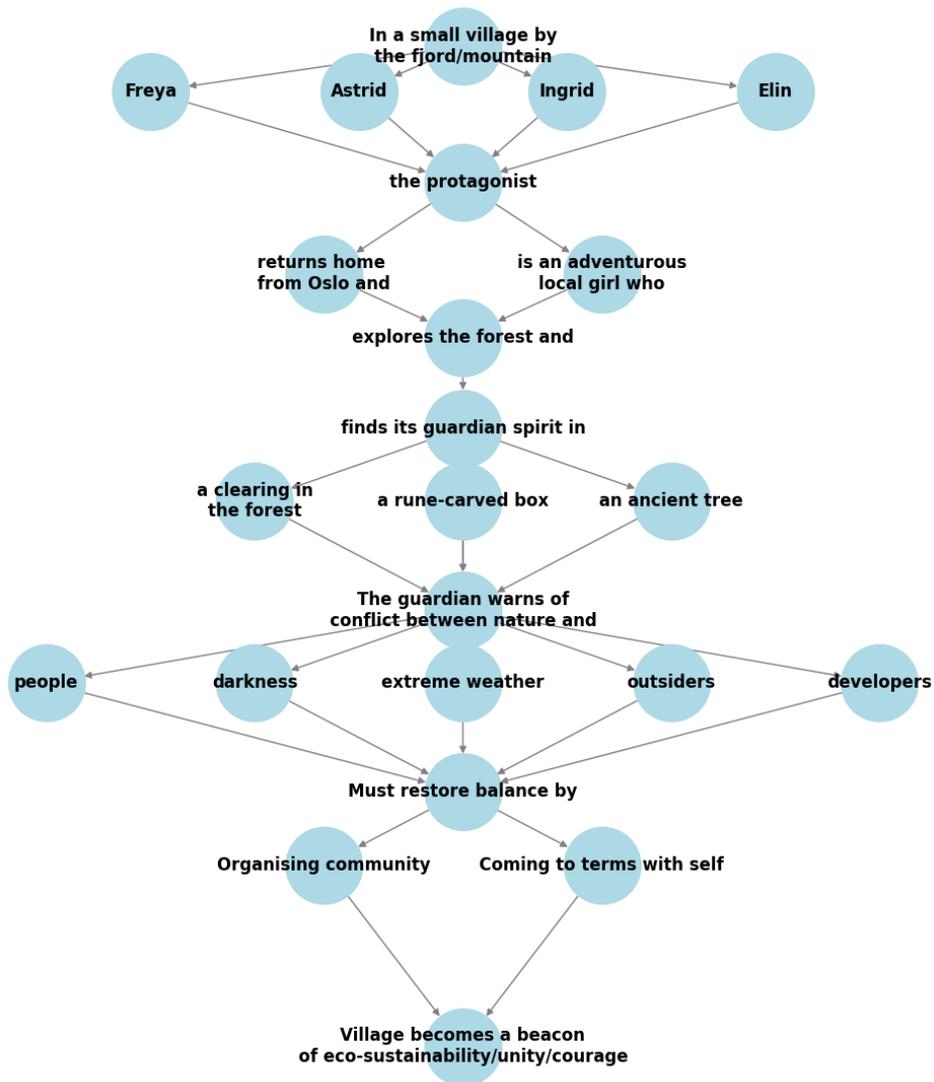

**Figure 6. Most of the Norwegian stories follow this structure.** This plot diagram was not generated computationally but by reading the stories and manually annotating them to identify shared plot points.

So where does the story come from? The basic structure of a small village or town, a protagonist often returning from the big city, and community organisation to solve an imbalance is very similar to the American stories, although the surface level has been changed to fit an idea of a "Norwegian" context: there are fjords, the names of the protagonists are stereotypically Norwegian. But many of these "Norwegian" stereotypes don't really hold up. Freya is a very rare name in contemporary Norway, and deep forests full of towering pines are not usually found next to fjords. Some of the stories have trials that the protagonist goes through that are reminiscent of video game mechanics, and perhaps the dramatic settings of these villages between fjords and mountains also "look like" video games.

One of the peculiarities of a language model is that it finds patterns in *all* its training data. So when it is asked to generate a story, it is not only drawing upon the *stories* in the training data, but upon all the texts. Perhaps the many online reviews of Freya birch spirit, an alcoholic beverage "budding with green, woody flavor", contribute to the strong presence of forests in gpt-4o-mini's synthetic imaginary of Norwegian stories and to Freya being such a common name for the protagonist in these stories. Freya, forests and fjords may cluster together and light up the same "neurons" in gpt-4o-mini's model, a cluster of neurons that is associated with "Norwegianness" and has nothing to do with actual Norwegian stories.

## Conclusion

Our analysis finds the same surface level stereotypes as other studies have found in AI-generated content, which is not surprising. A new finding is how similar plot structures are for stories across many countries. The stories in our dataset emphasise stability over change. Serious conflicts, such as war, are in the background rather than foregrounded. Conflict is resolved by the protagonist by leading community organisation.

This requires further research. Is the similarity caused by our prompt, or is it a clue to how LLMs are inferring narrative structures from the training data much as they infer grammatical rules allowing them to generate grammatically correct language?

There is a strong sense of nostalgia in the stories. This is particularly visible in the American stories, but an emphasis on heritage and tradition is present across the countries we looked at. Most of the stories feature a protagonist who revitalises the local community to bring back some of its lost glory. On the surface, this may sound similar to the generic fascist myth described by Roger Griffin, but it lacks both the revolutionary aspect of the fascist myth (nothing really changes in these stories) and the ultra-nationalism that Griffin's definition of fascism requires: ultra-nationalism is a type of nationalism that rejects liberal institutions and Enlightenment humanism (Griffin, 1991). These AI-generated stories express clichés of national identity, but the nation itself is never mentioned, and there is no drive to conquer or destroy other groups. The nation in these stories is more a stylistic flavour, a sprinkling of clichés, than it is a concrete reality. Regardless of the nationality given in the prompt, the stories share the setting of the generic "small town" strewn with a few national stereotypes.

The emphasis on stability rather than change is not surprising: LLMs identify correlations not causations (Chun, 2022; Rettberg, 2024). They work in the mathematical mode of $x = y$, stating that this is and this is and this is, rather than in the narrative mode of causality, where x causes y (Fletcher, 2021).

LLMs struggle with temporality. Everything is now. Or rather, time is just another feature, another parameter no more important than "Freya" being connected to "forests" and "fjords." This time-blindness causes an inability to portray historical events correctly. ChatGPT "lacks a proper temporal framework", "events often fluctuate within a broad time frame", and periodization is often incorrect, researchers have found (De Ninno & Lacriola, 2025, 195). While the stories in our dataset do have beginnings, middles and ends, often explicitly marked as such, there is no character development and little societal change other than the generic concluding statement that the village the story is set in becomes a beacon of hope and eco-sustainability.

Stories like these are being generated and posted online to earn money or for enjoyment and will in time become training data for new LLMs. The website Gathertales.com is an example: it promises "a rich variety of stories which come from various cultures and times", and allows readers to browse stories by country, genre or theme. Most of the stories appear to be fully AI-generated rather than retellings, and they tend to follow similar patterns to those analysed in this paper[8].

This technology isn't only being used to generate "AI slop", like the stories at Gathertales. It is being integrated into our everyday writing tools. As we type, our word processors and phones nudge us in particular directions, suggesting certain next words instead of others. Texts that are authored by humans alone may be revised by LLMs before they are published. A couple of weeks after we submitted this paper to *Open Research Europe*, we received it back with an email explaining that "We have undertaken a light text edit using the tool Paperpal Preflight; please check all edits to ensure you are happy with the changes made." The Word doc was a mess of hundreds of tiny changes. All the spelling and punctuation was Americanised. The wording of direct quotations was altered, the name "Rafael Pérez y Pérez" had been changed to "Rafael Pérez and Pérez", "finetuning the model" had become "refining the model" and "demonym" had become "denomination". These are all surface level alterations, although many of the AI-induced errors were culturally problematic, while others risked scientific integrity due to changing the meaning, or risked scientific misconduct when the text of a direct quotation was changed. However, Paperpal didn't attempt to change the structure of the text, as a human developmental editor might do. A future version of it might do so, perhaps suggesting that it be restructured to better fit a statistically calculated average genre of scientific paper. Our research is about AI-generated narratives, but AI will also shape other genres. *Open Research Europe* is certainly not the only journal or workplace to implement AI tools to support writing. This is happening throughout the public sector and in industry. We should be wary of implementing such tools without researching their impact.

---

[8] See for instance «The Witch Doctor of Otavalo», at https://www.gathertales.com/story/the-witch-doctor-of-otavalo/sid-1048

We hope our dataset can contribute to this research, both as a set of stories ready for computational or qualitative analysis and as an example that may inspire others to try different approaches, different prompts and different modes of analysis.

### Ethics and consent
Ethical approval and consent were not required.

### Data availability statement
DataverseNO. "A Dataset of 1500-Word Stories Generated by Gpt-4o-Mini for 252 Nationalities." https://doi.org/10.18710/VM2K4O

This project contains the following underlying data:

- 00_README.txt. (Description of the dataset).

- country_data.csv. (List of the countries we generated stories for with country names and demonyms, alpha-2 and alpha-3 country codes, emoji flags, regions, sub-regions and 2023 population numbers. See 00_README.txt for provenance. Note that we only generated stories for the 236 countries with demonyms).

- gpt-stories.zip (Zip file containing generated stories, summaries, word frequencies, sentiments and protagonist names for 236 countries, generated by gpt-4o-mini in January and February 2025).

Data is available under the terms of the CC0 1.0 licence.

### Software availability statement
Source code available from: https://github.com/AI-STORIES-ERC/GPT_stories

- Archived software available from: https://doi.org/10.5281/ZENODO.14939000

- License: MIT license